\pdfoutput=1

\documentclass[11pt]{article}

\usepackage{ACL2023}

\usepackage{amssymb}
\usepackage{times}
\usepackage{latexsym}
\usepackage{graphicx}
\usepackage{amsmath}
\usepackage{amsfonts}
\usepackage{multirow}
\usepackage{booktabs}
\usepackage{stmaryrd}
\usepackage{color}

\usepackage[T1]{fontenc}

\usepackage[utf8]{inputenc}

\usepackage{microtype}

\usepackage{inconsolata}

\title{TransESC: Smoothing Emotional Support Conversation via Turn-Level State Transition}

\author{Weixiang Zhao, Yanyan Zhao\thanks{\ \ Corresponding author} \ , Shilong Wang, Bing Qin \\
        Research Center for Social Computing and Information Retrieval\\
        Harbin Institute of Technology, China\\
        \texttt{\{wxzhao, yyzhao\}@ir.hit.edu.cn}}

\begin{document}
\maketitle
\begin{abstract}
Emotion Support Conversation (ESC) is an emerging and challenging task with the goal of reducing the emotional distress of people. Previous attempts fail to maintain smooth transitions between utterances in ESC because they ignore to grasp the fine-grained transition information at each dialogue turn. To solve this problem, we propose to take into account turn-level state \textbf{Trans}itions of \textbf{ESC} (\textbf{TransESC}) from three perspectives, including semantics transition, strategy transition and emotion transition, to drive the conversation in a smooth and natural way. Specifically, we construct the state transition graph with a two-step way, named transit-then-interact, to grasp such three types of turn-level transition information. Finally, they are injected into the transition-aware decoder to generate more engaging responses. Both automatic and human evaluations on the benchmark dataset demonstrate the superiority of TransESC to generate more smooth and effective supportive responses. Our source code is available at \url{https://github.com/circle-hit/TransESC}.
\end{abstract}

\section{Introduction}

Emotional Support Conversation (ESC) is a goal-directed task which aims at reducing individuals' emotional distress and bringing about modifications in the psychological states of them. It is a desirable and critical capacity that an engaging chatbot is expected to have and has potential applications in several areas such as mental health support, customer service platform, etc.

\begin{figure}[htbp]
\centering
\includegraphics[width=1.0\columnwidth]{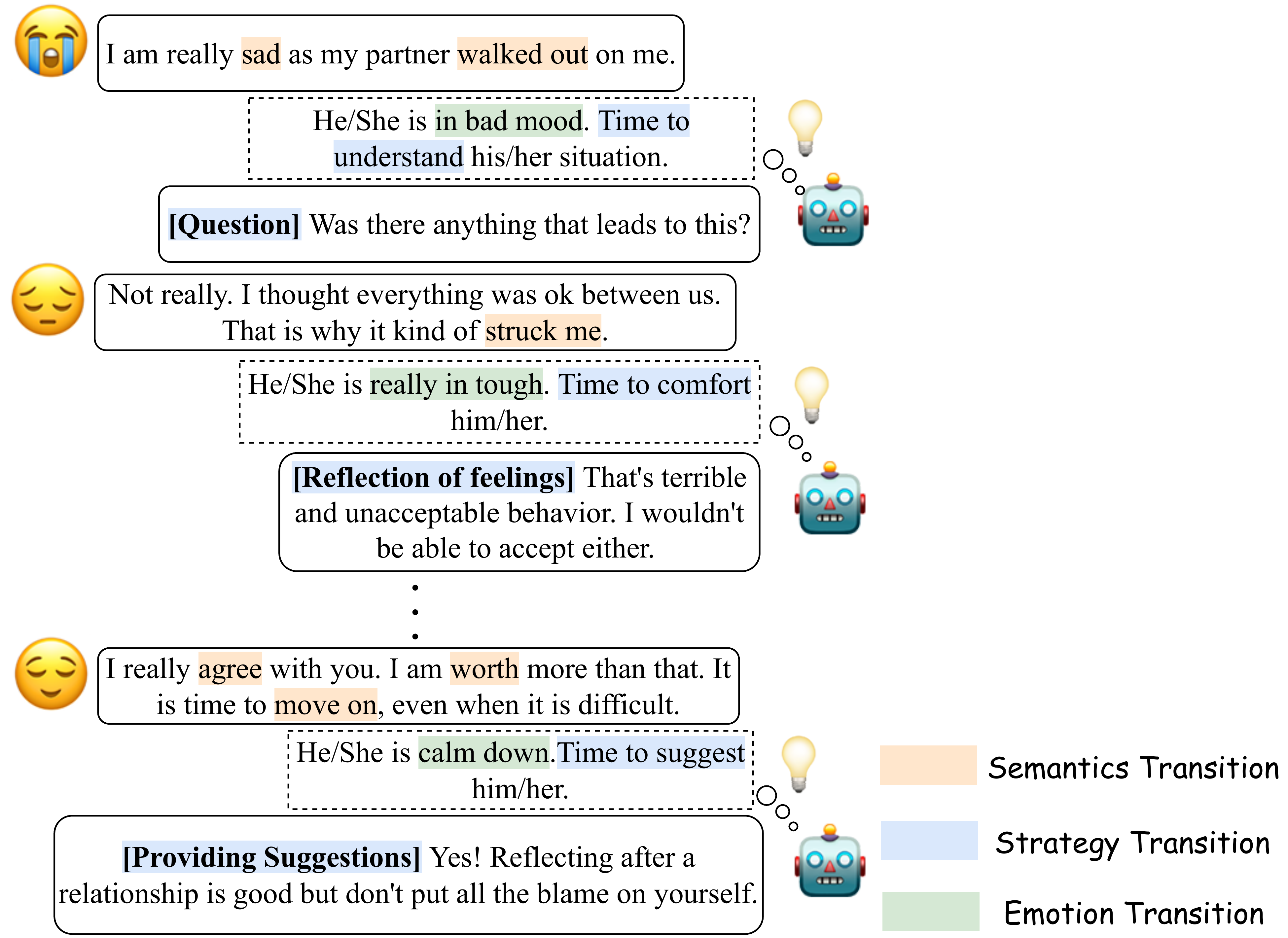}
\caption{An example for the turn-level state transitions during an emotional support conversation from the \textsc{ESConv} \citep{es} dataset.}
\label{example}
\end{figure}

Different from the emotional \citep{ecm} and empathetic \citep{ED_data} conversation, ESC is always of long turns, which requires skillful conversation procedures and support strategies to achieve the goal. For example, as shown in Figure \ref{example}, the supporter should firstly explore the situation to identify the problems faced by the seeker, and then try to comfort him. In the end, helpful suggestions are provided to help the seeker get rid of the tough. Intuitively, for such a complex and challenging task, a question is left: \emph{how to maintain smooth transitions between utterances from different procedures and drive the conversation in a natural way?} Previous works \citep{es,glhg,misc} fail to deal with this issue because they treat the dialogue history as a long sequence, ignoring to grasp the fine-grained transition information at each dialogue turn. We argue that considering such turn-level transition information plays the crucial role in achieving effective ESC, navigating the conversation towards the expected goal to reduce the seeker's distress in a smooth way. To achieve this, we model the transition information in ESC from three perspectives and refer to each one of them as a state.

\textbf{First}, it is a common phenomena that, even focusing on the same topic, the help seeker may tell different aspects or meanings as the conversation goes. We refer to it as \textbf{semantics transition} and take the example in Figure \ref{example}. To begin with, the help seeker feels sad to break up with the partner and does not know the reason (e.g. \emph{sad}, \emph{walked out}, \emph{struck me}). After receiving the warm and skillful emotional support from the supporter, he is relieved and encouraged to move forward (e.g. \emph{agree}, \emph{worth}, \emph{move on}). Thus, to fully comprehend the dialogue content with the goal of achieving effective emotional support, it is crucial to grasp such fine-grained semantic changes at each dialogue turn. 

\textbf{Second}, the timing to adopt proper support strategies constitutes another important aspect to achieve effective emotional support. In Figure \ref{example}, the supporter attempts to understand the seeker's problem via a \emph{Question} and comfort him by \emph{Reflection of feelings}. And the emotional support ends with the strategy \emph{Providing Suggestion} to help the seeker get through the tough. Such flexible combination and dependencies of different strategies forms the \textbf{strategy transition} in ESC, driving the conversation in the more natural and smooth way to solve the dilemma faced by the seeker.

\textbf{Finally}, it is also of vital importance to track the emotional state of the seeker as conversation develops. The seeker in Figure \ref{example} comes with a \emph{bad mood} and suffers from the \emph{tough} that his partner chooses to leave. As the ESC goes, his emotional state is changed and becomes \emph{calm down} to move on. Grasping such \textbf{emotion transition} can provide the supporter clear signals to apply proper strategies and offer immediate feedbacks to be aware of the effectiveness of the emotional support..

In this paper, in order to maintain smooth transitions between utterances in ESC and drive the conversation in a natural way, we propose to take into account turn-level state \textbf{Trans}itions of \textbf{ESC} (\textbf{TransESC}), including semantics transition, strategy transition and emotion transition. To be more specific, we construct the state transition graph for the process of emotional support. Each node consists of three types of states, representing semantics state, strategy state and emotion state of the seeker or the supporter at each dialogue turn. And seven types of edges form the path for information flow. Then we devise a two-step way, called transit-then-interact, to explicitly perform state transitions and update each node representation. During this process, ESC is smoothed through turn-level supervision signal that keywords of each utterance, adopted strategies by the support and immediate emotional states of the seeker are predicted by the corresponding state representations at each turn. Finally, we inject the obtained three transition information into the decoder to generate more engaging and effective supportive response.

The main contributions of this work are summarized as follows:
\begin{itemize}
    \item We propose to smooth emotional support conversation via turn-level state transitions, including semantics transition, strategy transition and emotion transition.
    \item We devise a novel model TransESC to explicitly transit, interact and inject the state transition information into the process of emotional support generation.
    \item Results of extensive experiments on the benchmark dataset demonstrate the effectiveness of TransESC to select the exact strategy and generate more natural and smooth responses.
\end{itemize}

\section{Related Works}

\subsection{Emotional Support Conversation}
\citet{es} propose the task of emotional support conversation and release the benchmark dataset \textsc{ESConv}. They append the support strategy as a special token into the beginning of each supportive response and the following generation process is conditioned on the predicted strategy token. \citet{glhg} propose a hierarchical graph network to utilize both the global emotion cause and the local user intention. Instead of using the single strategy to generate responses, \citet{misc} incorporate commonsense knowledge and mixed response strategy into emotional support conversation. More recently, \citet{cheng2022improving} propose look-ahead strategy planning to select strategies that can lead to the best long-term effects and \citet{peng2023fado} attempt to select an appropriate strategy with
the feedback of the seeker. However, all existing methods treat the dialogue history as a lengthy sequence and ignore the turn-level transition information that plays critical roles in driving the emotional support conversation in a more smooth and natural way.

\begin{figure*}[htbp]
\centering
\includegraphics[width=0.9\textwidth]{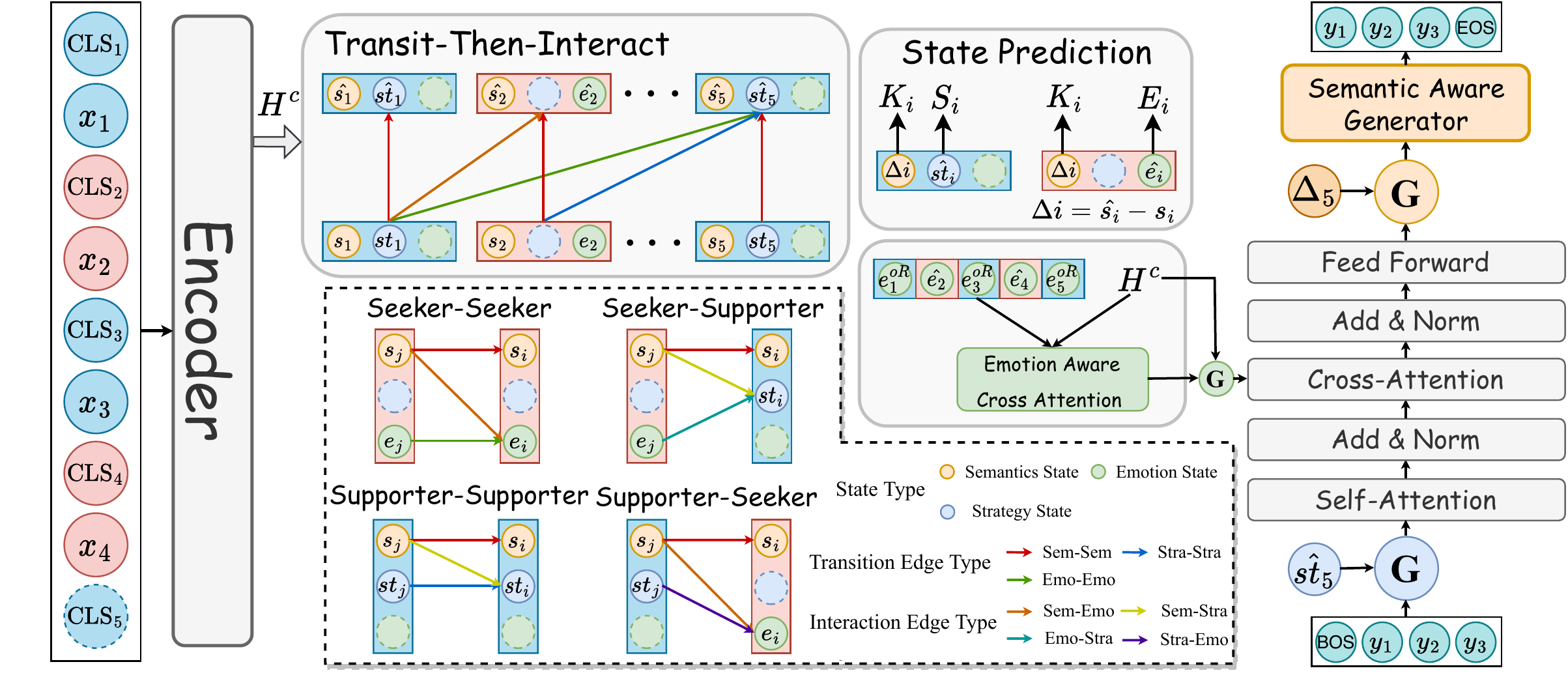}
\caption{The overall architecture of our proposed TransESC model, which mainly consists of three modules: Context Encoder, Turn-Level State Transition Module and Transition-Aware Decoder.}
\label{model}
\end{figure*}

\subsection{Emotional \& Empathetic Conversation}

Endowing emotion and empathy to the dialogue systems has gained more and more attentions recently. To achieve the former goal, both generation-based methods \citep{ecm,emoji,CDL} and retrieval-based \citep{QiuSLSL0020,LuTZ021} methods attempt to incorporate emotion into dialogue generation. However, it merely meets the basic quality of dialog systems. And to generate empathetic response, previous works incorporate affection \citep{alam2018annotating,ED_data,moel,mime,empdg,li2022knowledge}, cognition \citep{cem,zhao2022don} or persona \cite{zhong2020towards} aspects of empathy. Intuitively, expressing empathy is only one of the necessary steps to achieve effective emotional support. By contrast, emotional support is a more high-level ability that dialogue systems are expected to have.

\section{Preliminaries}
\subsection{ESConv Dataset}
Our research is carried out on the \textbf{E}motional \textbf{S}upport \textbf{Conv}ersation dataset, \textsc{ESConv} \citep{es}. In each conversation, the seeker with a bad emotional state seeks help to go through the tough. And the supporter is supposed to identify the problem that the seeker is facing, console the seeker, and then provide some suggestions to help the seeker to overcome their problems. The support strategies adopted by the supporter are annotated in the dataset and there are eight types of strategies (e.g., \emph{question}, \emph{reflection of feelings} and \emph{providing suggestions}). However, \textsc{ESConv} dataset does not contain keyword sets of each utterance and emotion labels \footnote{We use 6 emotion categories: joy, anger, sadness, fear, disgust, and neutral.} for the seeker's turn, we leverage external tools to automatically annotate them. More details about annotation are provided in Appendix \ref{app:ano}.

\subsection{Task Definition}
Formally, let $D=[X_1, X_2, \cdots, X_N]$ denotes a dialogue history with $N$ utterances between the seeker and the supporter, where the $i$-th utterance $X_i = [w^i_1, w^i_2 \cdots, w^i_m]$ is a sequence of $m$ words. And each utterance is provided with the extracted set of top $k$ keywords $K_i = [k^i_1, k^i_2 \cdots, k^i_k]$. Besides, the adopted support strategy $S_i$ of the supporter and the emotional state label $E_i$ of the seeker are also available for the turn-level supervision. The goal is to generate the next utterance $Y$ from the stand of the supporter that is coherent to the dialogue history $D$ and supportive to reduce the seeker's distress.

\section{Methodology}
The overall architecture of our proposed TransESC is shown in Figure \ref{model}. The dialogue representations are first obtained through context encoder. Then we grasp and propagate the fine-grained transition information, including semantics transition, strategy transition and emotion transition, in the Turn-Level State Transition Module. Finally, to generate more natural and smooth emotional support responses, such transition information is clearly injected into the Transition-Aware Decoder.

\subsection{Context Encoder}
We adopt Transformer encoder \citep{trans} to obtain the contextual representations of the dialogue history. Following previous works \citep{misc}, the dialogue is flattened into a word sequence. Then we append the special token $[\mathrm{CLS}]$ to the beginning of each utterance and another one for the upcoming response. And the context encoder produces the contextual embeddings $H^{c} \in \mathbb{R}^{N \times d_h}$.

\subsection{Turn-Level State Transition}
In this section, we propose to grasp the turn-level transition information, including semantics transition, strategy transition and emotion transition, to explicitly smooth the emotional support and drive the conversation in a natural way. Specifically, we construct the state transition graph, with three types of state for each node and seven types of edges, to propagate and update the transition information. And all the three states are supervised at each dialogue turn to predict the keyword set of each utterance, the adopted strategy of the supporter and the emotional state of the seeker.

\paragraph{State Transition Graph.} We construct the state transition graph to grasp and propagate transition information at each dialogue turn. To alleviate the impact of lengthy and redundant dialogue history, we perform the state transition within a fixed window size $w$. Specifically, we regard the current turn of supporter's response $u_e$ as the end and the $w$-th latest utterance $u_s$ spoken by the supporter as the start. All the utterances between $u_s$ and $u_e$ constitute the transition window.

\textbf{Nodes:} There are three types of states in total, making up each node in the transition graph. Since the adopted strategy and the emotional state are specified for the supporter and the seeker respectively, for the nodes from the supporter's turn, they include the semantics state and the strategy state, while the semantics state and the emotion state constitute the nodes for the seeker's turn.

\textbf{Edges:} We build edges to connect each node with all previous ones. Since there are two roles in ESC, it leads to four types of connection ways (e.g. Seeker-Seeker) between any two nodes. And seven types of edge are divided into two groups, the transition edges $\mathcal{T}$ and the interaction edges $\mathcal{I}$. For the former ones, they function to transit previous influences and grasp dependencies between states of the same type (e.g. Strategy-Strategy), while the later ones are devised to perform the interaction between different state types (e.g. Strategy-Emotion). The idea behind the interaction types is that decisions of the supporter to choose a certain strategy should focus on what the seeker said and are largely determined by emotional states of him/her. Also, what the supporter expressed and the adopted strategy could directly have impact on the emotional state of the seeker, leading the seeker into the better mood.

\paragraph{Graph Initialization.} Here we introduce the way to initialize three states for each node.

For the \textbf{semantics state} and the \textbf{strategy state} of each node, they are both initialized by the corresponding $[\mathrm{CLS_i}]$ token of each utterance.

And for the \textbf{emotion state}, in addition to initialized by the $[\mathrm{CLS_i}]$ token, we also leverage commonsense knowledge from the external knowledge base ATOMIC \citep{atomic} to imply the emotional knowledge of the seeker at each dialogue turn. Concretely, the generative commonsense transformer model COMET \citep{comet} is adopted to obtain the knowledge. We select relation type \textit{xReact} to manifest the emotional feelings of the seeker. Then the hidden state representations from the last layer of COMET are obtained as the emotional knowledge $csk_i$. The final representation of the emotion state is the sum of $[\mathrm{CLS_i}]$ and $csk_i$. Please refer to the Appendix \ref{app:csk} for the detailed implementation of COMET and definitions of the knowledge relation types in ATOMIC.

\paragraph{Transit-Then-Interact.} In order to explicitly grasp the turn-level transition information of the three states, we devise the two-step way Transit-Then-Interact (TTI) to propagate and update state representations of each node. Specifically, inspired by \citet{rmha}, the relation-enhanced multi-head attention (MHA) \citep{trans} is applied to update node representations from the information of the connected neighbourhoods. The formulation of vanilla MHA could be written as:
\begin{equation}\label{eq-encoder-multi-head}
\hat{v}_{i} = \underset{j \in \mathcal{N}}{\text{MHA}}(q_{i}, k_{j}, v_{j}),
\end{equation}
where $\text{MHA}(Q, K, V)$ follows the implementation of multi-head attention~\citep{trans}

And the key of relation-enhanced multi-head attention (R-MHA) is that we incorporate the embeddings of edge types into the query and the key. Thus, the two-step Transit-Then-Interact process operated on semantics states could be written as:
\begin{equation}\label{eq-encoder-multi-head}
s_{i}^{\prime} = \underset{e_{ij} \in \mathcal{T}}{\text{R-MHA}}(s_{i}+r_{ij}, s_{j}+r_{ij}, s_{j}),
\end{equation}
\begin{equation}\label{eq-encoder-multi-head}
s_{i}^{\prime \prime} = \underset{e_{ij} \in \mathcal{I}}{\text{R-MHA}}(s_{i}^{\prime}+r_{ij}, s_{j}^{\prime}+r_{ij}, s_{j}^{\prime}),
\end{equation}
where $e_{ij}$ is the edge type between the semantics states at $i$-th turn and that of $j$-th turn. $\mathcal{T}$ and $\mathcal{I}$ are the transition edge types and the interaction edge types, respectively. $r_{ij}$ is the embedding of $e_{ij}$. 

Then we dynamically fuse the results of transition $s_{i}^{\prime}$ and interaction $s_{i}^{\prime\prime}$ to obtain the updated semantics state $\hat{s}_i$:
\begin{equation}
\begin{aligned}
    \hat{s}_i &= g^{tti} \odot s_{i}^{\prime} + (1-g^{tti}) \odot s_{i}^{\prime\prime} \\
    g^{tti} &= \sigma([s_{i}^{\prime}; s_{i}^{\prime\prime}]W^{tti} + b^{tti}) \\
    \end{aligned}
\end{equation}
where $W^{tti} \in \mathbb{R}^{2d_h \times d_h}$ and $b^{tti} \in \mathbb{R}^{d_h}$ are trainable parameters.

Similarly, the ways to obtain the updated strategy state $\hat{st_i}$ and emotion state $\hat{e_i}$ are identical to that of the above semantics state $\hat{s_i}$.

\subsection{State Prediction}
We utilize the turn-level annotation to supervise the transition information, driving the emotional support conversation in a smooth and natural way. 
\paragraph{Semantic Keyword Prediction.} In order to measure the semantics transition more concretely, inspired by \citet{diaflow}, we calculate the difference $\Delta_i=\hat{s_i}-s_i$ between the semantics state before and after the operation TTI. Then we devise a bag-of-words loss to force $\Delta_i$ to predict the semantics keyword set $K_i = [k^i_1, k^i_2 \cdots, k^i_k]$ of the corresponding utterance. 
\begin{align}
    \mathcal{L}_{SEM} &= -\sum_{i=1}^{N}\sum_{j=1}^{k} \log p(k_{j}^{i}|\Delta_i) \nonumber \\ 
                      &= -\sum_{i=1}^{N}\sum_{j=1}^{k} \log f_{k_{j}^{i}}
\end{align}
where $f_{k_{j}^{i}}$ denotes the estimated probability of the $j$-th keyword $k_{j}^{i}$ in the utterance $u_i$. The function $f$ serves to predict the keyword set of the utterance $u_i$ in a non-autoregressive way:
\begin{equation}
    f = \textrm{softmax}(W^{sem}\Delta_i + b^{sem})
\end{equation}
where $W^{sem} \in \mathbb{R}^{d_h \times |V|}$, $b^{sem} \in \mathbb{R}^{|V|}$ and $V$ refers to the vocabulary size.

\paragraph{Supporter Strategy Prediction.} After the TTI module, we attempt to explicitly model the dependencies among the adopted supportive strategy during the ESC. Then we utilize the strategy label $S_i$ to specify the strategy state at each dialogue turn.
\begin{equation}
    \hat{y}_{str} = \textrm{softmax}(W^{str}\hat{st}_i + b^{str})
\end{equation}
where $\hat{y}_{str} \in \mathbb{R}^{n_s}$, $W^{str} \in \mathbb{R}^{d_h \times n_s}$ and $b^{sem} \in \mathbb{R}^{n_s}$. $n_s$ is the number of total available strategy.

Cross entropy loss is utilized and the loss function is defined as:
\begin{equation}
    \mathcal{L}_{STR} = - \frac{1}{N}\sum_{i=1}^N \sum_{j=1}^{n_s} \hat{y}_{str,i}^j \cdot log(y^j_{str,i})
\end{equation}
where $y^j_{str,i}$ stands for the ground-truth strategy label of the utterance $i$ from the supporter.

\paragraph{Seeker Emotion Prediction.} Similarly, the emotion states ${e}_i$ of each seeker's dialogue turn are also fed into another linear transformation layer:
\begin{equation}
    \hat{y}_{emo} = \textrm{softmax}(W^{emo}\hat{e}_i + b^{emo})
\end{equation}
where $\hat{y}_{emo} \in \mathbb{R}^{n_e}$, $W^{emo} \in \mathbb{R}^{d_h \times n_e}$ and $b^{emo} \in \mathbb{R}^{n_e}$. $n_e$ is the number of total available emotion.

Cross entropy loss is also utilized for training:
\begin{equation}
    \mathcal{L}_{EMO} = - \frac{1}{N}\sum_{i=1}^N \sum_{j=1}^{n_e} \hat{y}_{emo,i}^j \cdot log(y^j_{emo,i})
\end{equation}
where $y^j_{emo,i}$ is the ground-truth emotion label of the utterance $i$ from the seeker.

\subsection{Transition-Aware Decoder}
Finally, based on the vanilla Transformer decoder \citep{trans}, we devise the transition aware decoder to inject the turn-level transition information into the process of response generation.

 To make the generation process grounded on the selected strategy, we dynamically fuse the last strategy state $\hat{st}$ (the adopted strategy for the upcoming response) with the embeddings of the utterance sequence as the input of the decoder:
\begin{equation}
\begin{aligned}
    \hat{E_i} = g^{str} &\odot E_i + (1-g^{str}) \odot \hat{st} \\
    g^{str} &= \sigma([E_i; \hat{st}]W^1 + b^1) \\
    \end{aligned}
\end{equation}
where $W^1 \in \mathbb{R}^{2d_h \times d_h}$ and $b^1 \in \mathbb{R}^{d_h}$ are trainable parameters and $E_i$ is the $i$-th embedding token of the response.

And for the emotion transition information, we dynamically combine it with the output of the context encoder $H^c$ to explicitly incorporate the emotional states of the seeker. Specifically, the emotion states $e_i$ of the seeker and commonsense knowledge $e^{oR}_i$ of the supporter, which is generated by the COMET model under the relation type \emph{oReact} to imply what the emotional effect would exert on the seeker after the $i$-th utterance of the supporter, constitutes the emotional state sequence $H^{emo}$.
\begin{equation}
\begin{aligned}
    \hat{H} = g^{emo} & \odot H^c + (1-g^{emo}) \odot \hat{H}^{emo} \\
    \hat{H}^{emo} &= \textrm{Cross-Att}(H^c, H^{emo}) \\
    g^{emo} &= \sigma([H^c; \hat{H}^{emo}]W^2 + b^2) \\
    \end{aligned}
\end{equation}
where $W^2 \in \mathbb{R}^{2d_h \times d_h}$ and $b^2 \in \mathbb{R}^{d_h}$ are trainable parameters.

Thus, for the target response $Y=[y_1, y_2, \cdots, y_M]$, to generate the $t$-th token $y_t$, the hidden representation of it from the decoder can be obtained:
\begin{equation}
    h_t = \textrm{Decoder}(\hat{E}_{y < t}, \hat{H})
\end{equation}

In the end, we dynamically inject semantics transition information via the fusion of the last semantics difference representation $\Delta_i$ (latent semantic information for the upcoming utterance) and the hidden representation $h_t$ of the $t$-th token:
\begin{equation}
\begin{aligned}
    \hat{h} &= g^{sem} \odot h_t + (1-g^{sem}) \odot \Delta_i \\
    g^{sem} &= \sigma([h_t; \Delta_i]W^{sem} + b^{sem}) \\
    \end{aligned}
\end{equation}
where $W^{3} \in \mathbb{R}^{2d_h \times d_h}$ and $b^{3} \in \mathbb{R}^{d_h}$ are trainable parameters.

The distribution over the vocabulary for the $t$-th token can be obtained by a softmax layer:
\begin{equation}
    P(y_t \mid y_{<t}, D) = \textrm{softmax}(W\hat{h} + b)
\end{equation}
where $D$ is the input dialogue history.

We utilise the standard negative log-likelihood as the response generation loss function:
\begin{equation}
L_{gen}=-\sum_{t=1}^{M} \log P\left(y_{t} \mid D, y_{<t}\right).
\end{equation}

A multi-task learning framework is adopted to jointly minimize the response generation loss, the semantic keyword, strategy and emotion loss.
\begin{equation}
\mathcal{L}=\gamma_{1} \mathcal{L}_{GEN}+\gamma_{2} \mathcal{L}_{SEM}+\gamma_{3}\mathcal{L}_{STR} + \gamma_{4}\mathcal{L}_{EMO}
\end{equation}
where $\gamma_{1}$, $\gamma_{2}$, $\gamma_{3}$ and $\gamma_{4}$ are hyper-parameters.

\section{Experiments}

\subsection{Baselines}
We compare our proposed TransESC with the following competitive baselines. They are four empathetic response generators: \textbf{Transformer} \cite{trans}, \textbf{Multi-Task Transformer (Multi-TRS)} \cite{ED_data}, \textbf{MoEL} \cite{moel} and \textbf{MIME} \cite{mime}; and two state-of-the-art models on ESC task: \textbf{BlenderBot-Joint} \cite{es}, \textbf{GLHG} \cite{glhg} and \textbf{MISC} \cite{misc}. More details of them are described in Appendix \ref{app:bsl}.

\subsection{Implementation Details}
To be comparable with baselines, we implement our model based on BlenderBot-small \citep{blender} with the size of 90M parameters. The window size $w$ of turn-level transition is 2. The hidden dimension $d_h$ is set to 300 and the number of attention heads in relation enhanced multi-head attention and emotion aware attention graph are 16 and 4. Loss weights $\gamma_{1}$, $\gamma_{2}$, $\gamma_{3}$ and $\gamma_{4}$ are set to 1, 0.2, 1 and 1, respectively. AdamW \cite{adamw} optimizer with $\beta_1=0.9$ and $\beta_2=0.999$ is used for training. We vary the learning rate during the training process with the initial learning rate of 2e-5 and use a linear warmup with 120 warmup steps. And the training process is performed on one single NVIDIA Tesla A100 GPU with a mini-batch size of 20. For inference, following \citet{misc}, we also adopt the decoding algorithms of Top-$p$ and Top-$k$ sampling with $p$=0.3, $k$=30, temperature $\tau$=0.7 and the repetition penalty 1.03. 

\begin{table*}
\footnotesize
\centering
\begin{tabular}{cccccccccc}
\toprule
\textbf{Model} & \textbf{Acc} & \textbf{PPL} & \textbf{D-1} & \textbf{D-2} & \textbf{B-1} & \textbf{B-2} & \textbf{B-3} & \textbf{B-4} & \textbf{R-L} \\
\midrule
Transformer &- &89.61 &1.29 &6.91 &- &6.53 &- &1.37 &15.17\\
Multi-TRS   &- &89.52 &1.28 &7.12 &- &6.58 &- &1.47 &14.75\\
MoEL &- &133.13 &2.33 &15.26 &- &5.93 &- &1.22 &14.65 \\
MIME &- &47.51 &2.11 &10.94 &- &5.23 &- &1.17 &14.74\\
BlenderBot-Joint &17.69 &17.39 &2.96 &17.87 &18.78 &7.02 &3.20 &1.63 &14.92\\
GLHG &- &\textbf{15.67} &3.50 &\textbf{21.61} &\textbf{19.66} &7.57 &3.74 &2.13 &16.37 \\
MISC &31.67 &16.27 &4.62 &20.17 &16.31 &6.57 &3.26 &1.83 &17.24 \\
\midrule
TransESC (Ours) &\textbf{34.71} &15.85 &\textbf{4.73} &20.48 &17.92 &\textbf{7.64} &\textbf{4.01} &\textbf{2.43} &\textbf{17.51} \\
\bottomrule& 
\end{tabular}
\caption{Comparison of our model against state-of-the-art baselines in terms of the automatic evaluation. The best results among all models are highlighted in bold.}
\label{tab1}
\end{table*}

\begin{table}
\footnotesize
\centering
\resizebox{\linewidth}{!}{
\begin{tabular}{llcclcc}
\toprule
\multirow{2}{*}{\textbf{TransESC vs.}} & \multicolumn{3}{c}{\textbf{BlenderBot-Joint}} & \multicolumn{3}{c}{\textbf{MISC}}  \\
\cmidrule(r){2-4} \cmidrule(r){5-7}
                                  & Win  & Lose  & Tie & Win & Lose & Tie \\
    
\midrule
Fluency                & \textbf{54.7}$^\ddagger$   & 18.0             & 27.3            & \textbf{65.7}$^\ddagger$  &10.7  &23.7  \\
Identification           & \textbf{37.3}$^\ddagger$        & 16.0    & 46.7           &\textbf{32.0}   &19.3   &48.7  \\
Empathy               & \textbf{39.3}$^\ddagger$   & 7.0             & 53.7           &\textbf{48.0}$^\ddagger$  &5.7  &46.3  \\
Suggestion               & \textbf{37.0}   & 27.7             & 35.3           &\textbf{46.7}$^\dagger$   &17.3  &36.0  \\
\midrule
Overall                  &\textbf{51.7}$^\ddagger$   & 26.0             & 22.3           &\textbf{64.0}$^\ddagger$  &17.7  &18.3  \\
\bottomrule
\end{tabular}
}
\caption{The results of the human interaction evaluation (\%). TransESC performs better than all other models (sign test, $\ddagger$ / $\dagger$ represent \textit{p}-value < 0.05 / 0.1).}
\label{tab2}
\end{table}

\subsection{Evaluation Metrics}
\paragraph{Automatic Evaluation.} We apply four kinds of automatic metrics for evaluation: (1) Perplexity (\textbf{PPL}) measures the general quality of the generated responses; (2) BLEU-2 (\textbf{B-2}), BLEU-4 (\textbf{B-4}) \citep{bleu} and ROUGE-L (\textbf{R-L}) \citep{lin2004rouge} evaluate the lexical and semantic aspects of the generated responses; (3) Distinct-$n$ (\textbf{Dist}-$n$) \citep{dist} evaluates the diversity of the generated responses by measuring the ratio of unique $n$-grams; (4) Accuracy (\textbf{Acc}) of the strategy prediction is utilised to evaluate the model capability to choose the supportive strategy.

\paragraph{Human Evaluation.} Following \citet{es}, we recruit three professional annotators to interact with the models for human evaluation. Specifically, 100 dialogues from the test set of \textsc{ESConv} are randomly sampled. Then we ask the annotators to act as seekers under these dialogue scenarios and chat with the models. Given TransESC and a compared model, the annotators are required to choose which one performs better (or tie) following five aspects: (1) \textbf{Fluency}: which model generates more coherent and smooth responses; (2) \textbf{Identification}: which model explores the seeker's problems more effectively; (3) \textbf{Empathy}: which model is more empathetic to understanding the seeker's feelings and situations;  (4) \textbf{Suggestion}: which model offers more helpful suggestions; (5) \textbf{Overall}: which model provides more effective emotional support.

\section{Results and Analysis}
\subsection{Overall Results}
\begin{figure}[htbp]
\centering
\includegraphics[width=1.0\columnwidth]{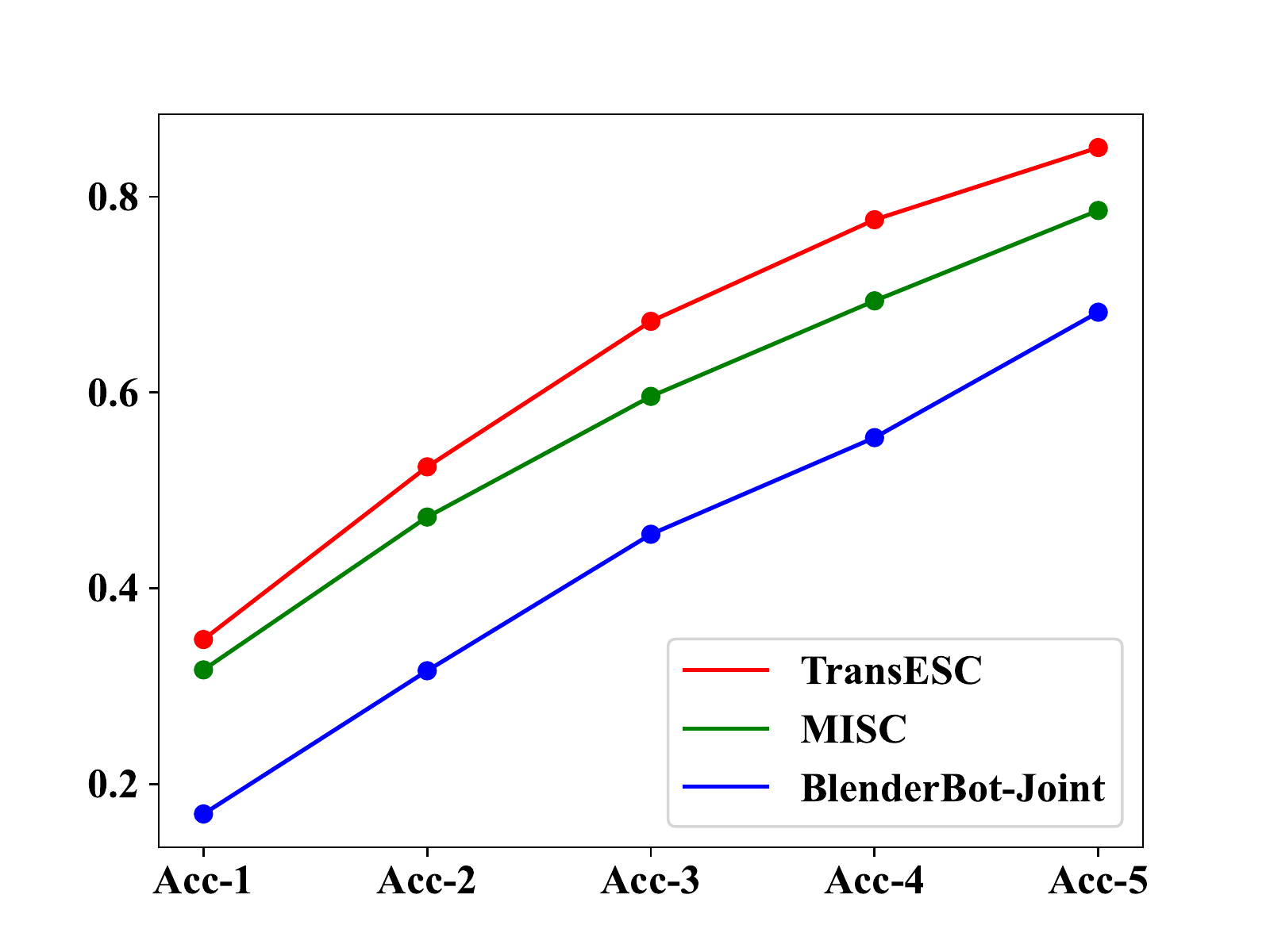}
\caption{The top-$n$ strategy prediction accuracy of TransESC and two baseline models.}
\label{acc}
\end{figure}

\paragraph{Automatic Evaluation.} As shown in Table~\ref{tab2}, TransESC achieves the new state-of-the-art automatic evaluation results. Benefiting from the grasp of three types of transition information in ESC, TransESC is capable of generating more natural and smooth emotional support responses in terms of almost all the metrics compared to the baselines. Compared with the empathetic response generators, the significant performance gain of TransESC demonstrates that eliciting empathy is only one of the critical procedures of ESC, while identifying the problems faced by the seeker and offering helpful suggestions also constitute the important aspects in ESC. Moreover, although the process of strategy prediction is also explored in BlenderBot-Joint and MISC, the prominent performance on strategy selection of TransESC can be ascribed to the explicit turn-level strategy transition modeling, which sufficiently capture the dependencies of different strategies adopted at each supporter's turn. As shown in Figure \ref{acc}, TransESC also outperforms baselines in terms of all the top-$n$ accuracy.

\paragraph{Human Evaluation.} For the evaluation setting, it is worth to mention that MISC takes the pre-conversation "situation" of the seeker as the input, which is not rational because the supporter can only comprehend what the seeker is facing as conversation goes. Thus, for the fair comparison, we do not input the "situation" for all three models. As shown in Table~\ref{tab2}, TransESC outperforms them in terms of all evaluation aspects. Specifically, it generates more fluent and smooth responses in terms of higher Fluency score, which verifies the benefits of incorporating turn-level transition information to maintain smooth transition between utterances. Also, although all three models may be comparable to identify problems of the seeker, TransESC could elicit more empathetic responses to comfort the seeker and then offer more helpful suggestions.

\begin{table}
\footnotesize
\centering
\begin{tabular}{lcccc}
\toprule
\textbf{Model} & \textbf{Dist-1} & \textbf{B-2} & \textbf{B-4} & \textbf{R-L}\\
\midrule
TransESC &4.73 &\textbf{7.64} &\textbf{2.43} &\textbf{17.51} \\
w/o Sem. Trans & 4.55 &7.04 &2.13 &17.37\\
w/o Stra. Trans &4.29 &6.68 &2.01 &17.15\\
w/o Emo. Trans &\textbf{4.82} &7.14 &2.22 &17.45\\
w/o T-L. Trans &4.19 &6.35 &1.94 &16.88 \\
\bottomrule
\end{tabular}
\caption{Results of ablation study. Sem./Stra./Emo./T-L. Trans refer to the semantics/strategy/emotion/all three types of turn-level transition, respectively.}
\label{tab3}
\end{table}

\begin{table*}
\small
\centering
\begin{tabular}{c|l}
\toprule
\textbf{Situation} & There is no hope, I am struggling with the pandemic and loneliness \\
\midrule
\multirow{4}{*}{\textbf{Context}} & \textbf{Supporter}: \textbf{[Affirmation and Reassurance]} I know that days can be really hard.  I think ... \\
 & \textbf{Seeker}: Yeah, I just kind of feel like a failure in life \\
 & \textbf{Seeker}: But I am trying, thanks. \\
 & \textbf{Supporter}: \textbf{[Affirmation and Reassurance]} I understand that there are things in your life ... \\
\midrule
\multirow{2}{*}{\textbf{BlenderBot-Joint}} & \textcolor{red}{\textbf{[Self-disclosure]}} I can understand why you are feeling this way. It is very difficult to see \\
& people be put down for the things that are bothering you. \\
\textbf{MISC} & \textcolor{red}{\textbf{[Others]}} I think you are doing the right thing! \\
\midrule
\multirow{2}{*}{\textbf{TransESC}} & \textcolor[rgb]{0, 0.7, 0}{\textbf{[Providing Suggestions]}} I think that you should try to focus on what is important to you. \\
& I know it can be hard to do that when you are feeling down but I believe that you can do it!\\
\midrule
\multirow{2}{*}{\textbf{Ground-Truth}} & \textbf{[Providing Suggestions]} When you feel up to it, do a search for temp agencies near you and \\
& hopefully they can give you some leads about a job. \\
\bottomrule
\end{tabular}

\caption{Case study of the generated supportive responses by our proposed TransESC and the baselines.}
\label{tab6}
\end{table*}


\begin{table}
\footnotesize
\centering
\begin{tabular}{ccccc}
\toprule
\textbf{Win. Size} & \textbf{Dist-1} & \textbf{B-2} & \textbf{B-4} & \textbf{R-L}\\
\midrule
$w=1$ &4.68 &7.49 &2.27 &17.25 \\
$w=2$ & \textbf{4.73} &\textbf{7.64} &\textbf{2.43} &\textbf{17.51}\\
$w=3$ & 4.49 &6.52 &2.26 &17.29\\
$w=4$ & 4.39 &7.04 &2.12 &17.29\\
$w=5$ & 4.71 &6.98 &2.17 &17.24\\
\bottomrule
\end{tabular}
\caption{Results of our proposed model with different lengths of transition window $w$.}
\label{tab4}
\end{table}

\subsection{Ablation Study}
To explore the impact of three types of transition information, we remove the corresponding state representation with edges in the transition graph, the turn-level label prediction and the injection into the decoder. Besides, to explore the effect of turn-level transition process, we also discard it by predicting three states with the whole dialogue history.

As shown in Table \ref{tab3}, the ablation of any types of transition information can lead to a drop in the automatic evaluation results, demonstrating the effectiveness of each one of them. To be more specific, the ablation of the strategy transition (w/o Stra.Trans) causes the most significant performance drop. The reason is that selecting the proper strategy to support the seeker plays the most pivotal role in ESC. And the impact of emotion transition (w/o Emo.Trans) is relatively small. It may be attributed to the noise of annotated emotion labels and the generated emotional knowledge. 

Moreover, when we remove the whole process of turn-level state transition, the significant performance drop verifies our contribution that grasping the fine-grained transition information can drive the ESC in a more smooth and natural way.

\subsection{Case Study}
In Table~\ref{tab6}, we show a case with responses generated by TransESC and two baselines. With the emotion transition and strategy transition, after several turns of comforting, TransESC senses the emotion state \emph{joy} of the seeker and it is 
time to offer helpful suggestions with the correct predicted strategy. And through semantics transition, it grasp the determination of the seeker to suggest him to have a try and encourage him to face the failure. By contrast, MISC and BlenderBot-Joint drive the conversation improperly, leading to the ineffective responses.

\subsection{Length of Transition Window}
We adjust different lengths of transition window for a deeper analysis of the impact of transition information modeling. Results are shown in Table \ref{tab4}. The model with the transition window length of 2 achieves the best performance. On the one hand, capturing the transition information in the shorter window could not sufficiently comprehend dependencies of utterance transition in the dialogue history. On the other hand, much more redundant transition information may be incorporated by the model with longer transition window, which would weaken the performance of our model.

\section{Conclusion and Future Work}
In this paper, we propose TransESC to generate emotional support via turn-level state transition information incorporated, including semantics transition, strategy transition and emotion transition. We construct the transition graph with the two-step way, transit-then-interact, to grasp and supervise the transition information at each dialogue turn. Experimental results on both automatic and human evaluation demonstrate the superiority of TransESC to generate more smooth responses.

In the future, we will explore more characteristics in ESC such as persona to generate more natural responses.

\section{Limitations}
Although our proposed method exhibits great performance to generate more smooth and natural emotional support than baseline models,  we argue that the research on this field still has a long way to go. We conclude three aspects that may inspire further exploration. First, the automatically annotated emotion labels may be a little bit coarse and may not accurately manifest the emotional states of the seeker. Second, since various types of commonsense knowledge are not introduced, the current chatbots always generate general and safe responses, failing to provide specific and personalized suggestions to help the seeker get over the dilemma. Finally, current automatic evaluation metrics are still not rational and proper to measure the ability of chabots to provide emotional support. It is desirable to build better evaluation metrics for this.

\section{Ethics Statement}
The open-source benchmark dataset \textsc{ESConv} \citep{es} used in our experiments is well-established and collected by employed crowd-sourced workers, with user privacy protected and no personal information involved. And for our human evaluation, all participants are volunteered and transparently informed of our research intent, with reasonable wages paid.

Moreover, our research only focuses on building emotional support systems in daily conversations, like the one to seek the emotional support from our friends or families. It is worth to mention that we do not claim to construct chatbots that can provide professional psycho-counseling or professional diagnosis. This requires particular caution and further efforts to construct a safer emotional support system, which is capable of detecting users who have tendencies of self-harming or suicide.

\section*{Acknowledgements}

We thank the anonymous reviewers for their insightful comments and suggestions. This work was supported by the National Key RD Program of China via grant  2021YFF0901602 and the National Natural Science Foundation of China (NSFC) via grant 62176078.

\bibliography{anthology,custom}
\bibliographystyle{acl_natbib}

\newpage

\appendix

\begin{table}[htbp]
\centering
\small
\begin{tabular}{lccc} 
\toprule
\textbf{Category}& \textbf{Train} & \textbf{Dev} & \textbf{Test}\\
\midrule
\# dialogues & 14116& 1763&1763 \\
Avg. \# words per utterance &18.16 &18.01 &18.01 \\
Avg. \# turns per dialogue &8.61 &8.58 &8.48 \\
Avg. \# words per dialogue &156.29 &154.58 &152.79 \\
\bottomrule
\end{tabular}
\caption{The statistics of processed ESConv dataset.}
\label{tab:1}
\end{table}

\section{\textsc{ESConv} Dataset}
\label{app:ano}

\subsection{Keyword and Emotion Annotation}
Since the original \textsc{ESConv} dataset does not contain keyword sets of each utterance and emotion labels for the seeker's turn, we leverage external tools to annotate them. To obtain the keyword set of each utterance, we use TF-IDF method. The vocabulary and IDF term are learned from the training set of \textsc{ESConv}. Then for each utterance, we apply TF-IDF to obtain the top $k$ keywords.

For the emotion labels, we fine-tune the BERT model \citep{bert} on a fine-grained emotion classification dataset, GoEmotions \citep{goemotion}. The the finetuned BERT model achieve an accuracy of 71\% on test set, indicating that it is reliable for emotion classification. Then it is used to annotate an emotion label for each utterance from the seeker's turn.

\subsection{Dataset Statistics}
We carry out the experiments on the dataset \textsc{ESConv}~\citep{es} \footnote{https://github.com/thu-coai/Emotional-Support-Conversation}. For pre-processing, following \citep{misc} we truncate the conversation examples every 10 utterances, and randomly spilt the dataset into train/valid/test set with the ratio of 8:1:1. The statistics is given in Table~\ref{tab:1}.

\subsection{Definitions of Strategies}
There are overall 8 types of support strategies that are originally annotated in the \textsc{ESConv} dataset:
\begin{itemize}
    \item \textbf{Question}: ask for information related to the problem to help the help-seeker articulate the issues that they face.  
    \item \textbf{Restatement or Paraphrasing}: 
    a simple, more concise rephrasing of the support-seeker’s statements that could help them see their situation more clearly.
    \item \textbf{Reflection of Feelings}: describe the help-seeker’s feelings to show the understanding of the situation and empathy.
    \item \textbf{Self-disclosure}: share similar experiences or emotions that the supporter has also experienced to express your empathy.
    \item \textbf{Affirmation and Reassurance}: affirm the help-seeker’s ideas, motivations, and strengths to give reassurance and encouragement. 
    \item \textbf{Providing Suggestions}: provide suggestions about how to get over the tough and change the current situation.
    \item \textbf{Information}: provide useful information to the help-seeker, for example with data, facts, opinions, resources, or by answering questions.
    \item \textbf{Others}: other support strategies that do not fall into the above categories.
\end{itemize}

\section{Commonsense Knowledge Acquisition}
\label{app:csk}

\subsection{Description of ATOMIC Relations}
ATOMIC \citep{atomic} is an atlas of everyday commonsense reasoning and organized through textual descriptions of inferential knowledge, where nine if-then relation types are proposed to distinguish causes vs. effects, agents vs. themes, voluntary vs. involuntary events, and actions vs. mental states. We give the brief definition of each relation.
\begin{itemize}
    \item \textbf{xIntent}: Why does PersonX cause the event?
    \item \textbf{xNeed}: What does PersonX need to do before the event?
    \item \textbf{xAttr}: How would PersonX be described?
    \item \textbf{xEffect}: What effects does the event have on PersonX?
    \item \textbf{xWant}: What would PersonX likely want to do after the event?
    \item \textbf{xReact}: How does PersonX feel after the event?
    \item \textbf{oReact} How does others' feel after the event?
    \item \textbf{oWant} What would others likely want to do after the event?
    \item \textbf{oEffect} What effects does the event have on others?
\end{itemize}

\subsection{Implementation Details of COMET}
The generative commonsense transformer model COMET \citep{comet} is adopted to obtain the knowledge. We select relation types \textit{xReact} to manifest the emotional feelings of the seeker at each dialogue turn. Specifically, we adopt the BART-based \citep{bart} variation of COMET, which is trained on the ATOMIC-2020 dataset \citep{atomic2020}. And given each utterance $X_i$ belonging to the self to form the input format $(X_i, r, [\textrm{GEN}])$, COMET would generate descriptions of inferential content under the relation $r$. Then the hidden state representations from the last layer of COMET are obtained as knowledge representation.

\section{Baselines}
\label{app:bsl}
\begin{itemize}
    \item \textbf{Transformer} \cite{trans}: The vanilla Transformer-based encoder-decoder generation model.
    \item \textbf{Multi-Task Transformer (Multi-TRS)} \cite{ED_data}: A variation of the vanilla Transformer with an auxiliary task to perform emotion perception of the user.
    \item \textbf{MoEL} \cite{moel}: A Transformer-based model that captures emotions of the other and generates an emotion distribution with multi decoders. Each decoder is optimized to deal with certain emotions and generate an empathetic response through softly combining the output emotion distribution.
    \item \textbf{MIME} \cite{mime}: Another Transformer-based model with the notion of mimicing the emotion of the other to a varying degree by group emotions into two clusters. It also introduces stochasticity to yield emotionally more varied empathetic responses.
    \item \textbf{BlenderBot-Joint} \cite{es}: A strong baseline model on the \textsc{ESConv} dataset, which prepends the special strategy token at the beginning of responses and conditions the generation process on it.
    \item \textbf{GLHG} \cite{glhg}: A hierarchical graph neural network to model the relationships between the global user’s emotion causes and the local intentions for emotional support dialogue generation.
    \item \textbf{MISC} \cite{misc}: An encoder-decoder model that leverages external commonsense knowledge to infer the seeker’s fine-grained emotional status and respond skillfully using a mixture of strategy.
\end{itemize}

\end{document}